\documentclass[10pt, a4paper, logo]{googledeepmind}
\usepackage{times}

\usepackage{hyperref}
\usepackage[leftcaption]{sidecap} % caption 在右边

\usepackage{amsmath,amsfonts,bm}
\usepackage{natbib}
\usepackage{fancyhdr}
\def\eqref#1{equation~\ref{#1}}
\def\1{\bm{1}}

\DeclareMathAlphabet{\mathsfit}{\encodingdefault}{\sfdefault}{m}{sl}
\SetMathAlphabet{\mathsfit}{bold}{\encodingdefault}{\sfdefault}{bx}{n}

\newcommand{\E}{\mathbb{E}}

\usepackage{hyperref}
\usepackage{url}
\usepackage{booktabs}
\usepackage{graphicx}
\usepackage{subcaption}
\usepackage{amssymb}
\usepackage[most]{tcolorbox}
\usepackage[table]{xcolor}
\tcbuselibrary{breakable} % 加载分页支持库
\usepackage{amsmath}
\usepackage{siunitx}
\usepackage{colortbl}

\usepackage{microtype}
\usepackage{color}
\usepackage{wrapfig}
\usepackage{natbib}
\usepackage{colortbl}
\usepackage{microtype}
\usepackage{graphicx}
\usepackage{booktabs} % for professional tables
\usepackage{array}
\usepackage{textcomp}
\usepackage{stfloats}
\usepackage{float}
\usepackage{verbatim}
\usepackage[ruled, linesnumbered, lined]{algorithm2e}
\usepackage{multirow}
\usepackage{enumitem}

\definecolor{BestColor}{HTML}{C8E6C9}  % 一个柔和的绿色
\definecolor{SecondBestColor}{HTML}{FFF9C4} % 一个非常淡的黄色

\usepackage{tcolorbox}
\usepackage{amsmath,amsfonts}

\definecolor{ggg}{RGB}{26,179,0}
\definecolor{rrr}{RGB}{179,0,0}
\definecolor{oodc}{RGB}{31,73,121}
\definecolor{idc}{RGB}{68,142,68}

\definecolor{mygray}{gray}{0.9}

\def\Bias#1#2{\bm{b}}
\newtcolorbox{examplebox}[2][]{ % 允许传入可选参数 [#1] 和必选标题参数 {#2}
    breakable, % 关键：允许跨页分割
    enhanced, % 增强模式（可选，支持更多样式）
    colback=white, % 框体内背景色
    colframe=cyan, % 边框颜色
    coltitle=white, % 标题文字颜色
    fonttitle=\bfseries, % 标题字体加粗
    title=#2, % 框体标题（第二个必选参数）
    overlay middle={\draw[cyan, line width=1pt](frame.south west)--(frame.south east);}, % 分割处添加横线
    overlay last={\draw[cyan, line width=1pt](frame.south west)--(frame.south east);}, % 最后一页底部横线
    #1 % 允许在调用时传入其他可选参数以覆盖默认样式
}

\usepackage[T1]{fontenc}
\usepackage{booktabs}      % 用于漂亮的表格线 (toprule, midrule, etc.)
\usepackage{graphicx}      % 用于 \resizebox
\usepackage[table]{xcolor} % 用于颜色
\usepackage{siunitx}       % 用于 S 列，实现小数点对齐
\usepackage{etoolbox}      % 用于 \ifstrequal，实现条件判断
\usepackage[normalem]{ulem}     % 用于更灵活的下划线 \uline

\definecolor{impcolor}{HTML}{2E8B57} % 提升使用的海绿色 (SeaGreen)

\newcommand{\improvementstyle}[1]{$^{\textcolor{impcolor}{\tiny #1}}$}

\newcommand{\scoreimp}[2]{%
  \textbf{#1}%
  \ifstrequal{#2}{+0.0}{}{%
    \ifstrequal{#2}{0.0}{}{%
      \makebox[0pt][l]{\improvementstyle{#2}}%
    }%
  }%
}

\definecolor{smopdrow}{gray}{0.82}
\definecolor{defaultrow}{gray}{0.93}
\newcommand{\cL}{\mathcal{L}}

\providecommand{\E}{}\renewcommand{\E}{\mathbb{E}}

\title{SMOPD: Multi-Reward Reinforcement Learning\\via \textbf{S}pecialize-and-\textbf{M}erge \textbf{O}nline \textbf{P}olicy \textbf{D}istillation}

\author[1,2]{Wen Wang\textsuperscript{\ddag}}
\author[1]{Jiahua Bao\textsuperscript{\ddag}}
\author[1]{Tu Yongsiqi\textsuperscript{\ddag}}
\author[1]{Yihao Liu\textsuperscript{\ddag}}
\author[1]{Haotian Zhou\textsuperscript{\ddag}}
\author[1]{Haoxuan Ma\textsuperscript{\ddag}}
\author[1]{Mengyu Zhou\textsuperscript{\dag}}
\author[1]{Wenkui Fan\textsuperscript{\ddag}}
\author[2]{Junwei He}
\author[1]{Xiaoxi Jiang}
\author[1]{Guanjun Jiang}
\affil[1]{Qwen Large Model Application Team, Alibaba}
\affil[2]{University of Chinese Academy of Sciences}

\correspondingauthor{zhoumengyu.zmy@alibaba-inc.com}

\begin{abstract}
We aim to improve model performance in multi-reward reinforcement learning training process.
Existing Group reward-Decoupled Normalization Policy Optimization (GDPO) has mitigated the issue of reward signals masking one another during direct scalarization by normalizing each reward dimension separately before aggregation.
However, our experiments show that GDPO still struggles to balance reward signals with different granularities.
Specifically, in some particular training tasks, the model may receive a dense reward that assigns fine-grained scores ranging from 0.1 to 1.0, together with a sparse reward that provides only binary feedback of either 0 or 1.
In such cases, we find that the sparse reward may provide an insufficient optimization signal, preventing its corresponding capability from being effectively reinforced.
Therefore,
\textbf{\emph{how can we strengthen the optimization signal from the sparse reward without sacrificing the capability already learned from the fine-grained reward?}}
To overcome this limitation, we propose \textbf{Specialize-and-Merge Online Policy Distillation (SMOPD)}, a two-stage training method for multi-reward optimization.
\textbf{\emph{Stage1-Specialize}}: SMOPD first employs reward-priority configurations to train multiple reward-specialized teachers, allowing each reward to be learned under conditions where its signal can effectively drive optimization.
\textbf{\emph{Stage2-Merge}}: SMOPD then utilizes online policy distillation to combine the reward-specialized capabilities of these teachers into a single student policy, while maintaining balanced task-level optimization.
To validate our method, we conduct experiments on two multi-reward settings: complementary rewards (tool-calling accuracy and format) and conflicting rewards (helpful and harmless rewards).
Based on above settings, SMOPD outperforms GDPO across 1.5B, 3B and 7B backbones.
\end{abstract}

\begin{document}
\maketitle

\section{Introduction}

Reinforcement learning from human feedback (RLHF) has become the dominant paradigm for aligning large language models with human preferences \cite{ouyang2022instructgpt,christiano2017rlhf}.
In multi-reward alignment, prior work commonly adopts GRPO \cite{shao2024deepseekmath} to optimize a single policy with several reward signals.
However, because GRPO first sums different rewards and then computes group-relative advantages, reward dimensions with different scales or reward combinations can mask one another during scalarization, causing the final advantage to lose reward-specific information.
GDPO \cite{liu2026gdpo} mitigates this aggregation-level issue by normalizing each reward dimension separately before aggregation, thereby preserving the contribution of each reward in the training signal.
Building on this reward-level decomposition, GD$^2$PO \cite{liu2026gd2po} further addresses conflicts among reward dimensions by filtering rollouts with severe reward-wise disagreement and reweighting queries according to reward consensus.
Despite these improvements, GDPO and GD$^2$PO still focus on how observed reward signals are normalized and combined within a single policy.

\begin{wrapfigure}{r}{0.5\textwidth}
\vspace{-12pt}
\centering
\captionsetup{font=footnotesize}
\includegraphics[width=.95\linewidth]{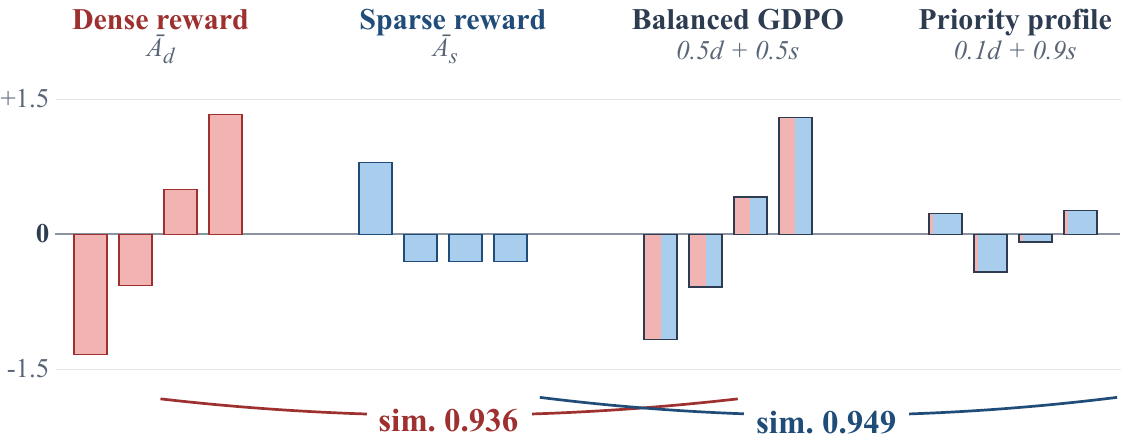}
\centerline{\footnotesize (a) Reward-density effect.}
\vspace{3pt}
\includegraphics[width=.75\linewidth]{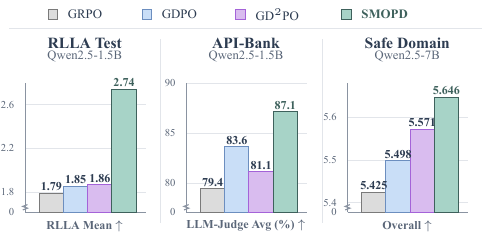}
\centerline{\footnotesize (b) Performance comparison.}
\caption{
\textbf{(a)} In a batch of $8$ prompt groups with $4$ rollouts each, the dense reward vector is $[0,1,2,3]$ in all $8$ groups, whereas the sparse reward vector is $[0,0,0,0]$ in $7$ groups and $[1,0,0,0]$ in the remaining group. The bars show batch-averaged advantage profiles under balanced weights $(0.5,0.5)$ and sparse-priority weights $(0.1,0.9)$.
\textbf{(b)} Main results comparing SMOPD with multi-reward baselines on RLLA Test (1.5B), API-Bank LLM-Judge (1.5B), and Safe Domain (7B).
}
\label{fig:smopd_teaser}
\vspace{-10pt}
\end{wrapfigure}

This leaves open a different problem: \textbf{Reward dimensions can differ not only in scale, but also in how often they provide informative learning signals.}
Group-based advantage estimation relies on reward differences among rollouts sampled for the same prompt.
Dense reward distributions can rank sampled responses in most rollout groups, continuously providing reliable optimization signals.
Sparse reward distributions, by contrast, may assign the same value to most responses in a group, leaving that reward dimension with little useful within-group variation.
In such groups, per-reward normalization cannot create an informative signal where no response-level distinction exists.
As a result, under balanced priorities, training is still dominated by dense reward dimensions, while the occasional signal from sparse reward distributions can be overwhelmed before it accumulates.
Figure~\ref{fig:smopd_teaser}(a) illustrates that the balanced GDPO advantage remains closely aligned with the dense reward.%, whereas a sparse-reward priority profile recovers the sparse direction.

Therefore, single-policy multi-reward RL faces an inherent tension: Raising the priority of a sparse reward to make it learnable inevitably skews the final balance, yet keeping a balanced priority leaves it overshadowed.
Our experiments confirm this issue in tool calling \cite{schick2023toolformer,qin2024toolllm}, where a dense accuracy reward is paired with a sparse binary format reward.
Under balanced priorities, GDPO behaves similarly to GRPO, with format compliance remaining below $9\%$.
When the format reward is made dominant, the model reliably learns the required output structure, but the resulting policy is optimized under a deliberately skewed objective.
This suggests that GDPO alleviates aggregation-level reward masking, yet still struggles with the signal-density imbalance between sparse and dense reward distributions.

\begin{figure*}[t]
\centering
\includegraphics[width=\textwidth]{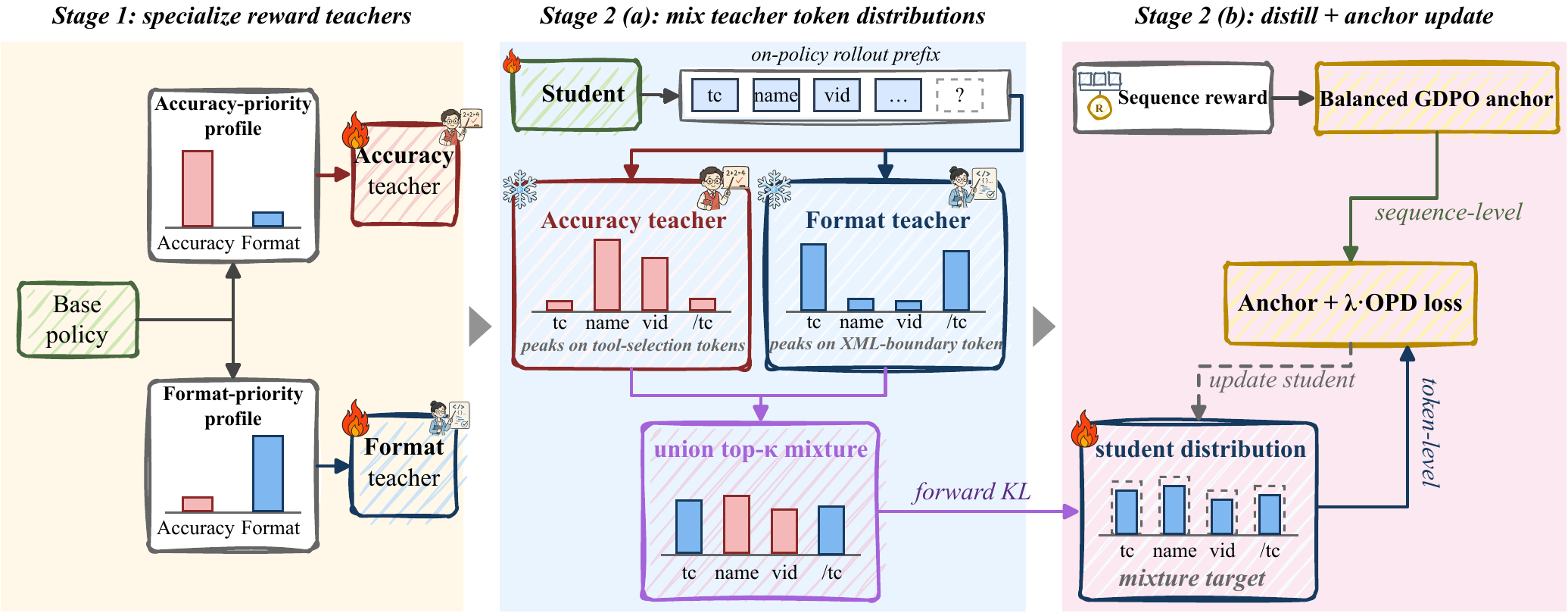}
\caption{Overview of SMOPD. \emph{Stage~1}: complementary GDPO priority profiles produce an accuracy teacher and a format teacher from the same base policy. \emph{Stage~2 (a)}: on the student's rollout prefix, frozen teachers emit top-$\kappa$ next-token distributions. The accuracy teacher peaks on tool-selection tokens, the format teacher peaks on XML boundary tokens, and the union-top-$\kappa$ mixture retains both. \emph{Stage~2 (b)}: the student is updated through two-level signals. Token-level OPD fits its distribution to the teacher mixture via forward KL, while the sequence-level task anchor optimizes the balanced multi-reward objective.}
\label{fig:mopd_framework}
\end{figure*}

To resolve this tension, we propose \textbf{Specialize-and-Merge Online Policy Distillation (SMOPD)}, a two-stage method for multi-reward optimization.
SMOPD is designed to preserve the strengths learned under dense reward distributions while improving the model's sensitivity to sparse reward distributions.
\textbf{\emph{Stage1-Specialize}}.
SMOPD uses reward-priority configurations to train multiple reward-specialized teachers from the same base policy.
Instead of balancing all rewards within a single policy, we assign each teacher a reward-priority profile that amplifies its target reward in the optimization signal. In particular, assigning higher priority to a sparse reward counteracts its weaker and less frequent learning signal, enabling the policy to more effectively capture the optimization direction induced by that reward. Figure~\ref{fig:smopd_teaser}(a) illustrates: a sparse-reward priority profile recovers the sparse reward direction.
This allows sparse or hard-to-learn reward distributions to be acquired in a favorable regime, while capabilities learned from dense reward distributions are preserved by complementary teachers.
\textbf{\emph{Stage2-Merge}}.
SMOPD merges these reward-specialized teachers into a single student policy through online policy distillation.
The student learns from the teachers' reward-specialized behaviors on its own rollouts, while a balanced GDPO anchor maintains task-level optimization over the original multi-reward objective.
In this way, SMOPD first acquires different reward strengths through specialization and then balances them through policy-level merging, producing one unified policy instead of a set of separate specialists.
Figure~\ref{fig:mopd_framework} summarizes the full workflow.

Experimentally, we validate SMOPD across model scales across 1.5B, 3B and 7B with different model families(Qwen2.5 \cite{qwen2024qwen25} and Llama-3.2 \cite{grattafiori2024llama3}) under two reward structures, including complementary rewards (tool-calling accuracy and format) and conflicting rewards(helpful and harmless rewards) \cite{bai2022hh,bai2022constitutional}.
Our teacher analysis further surfaces structure that a scalarized objective cannot represent.

Our contributions are:

\begin{itemize}
    \item \textbf{SMOPD resolves the reward-balancing tension in multi-reward RL.}
    Sparse reward distributions are often learnable only when made dominant, but such skewed priorities sacrifice other objectives in a single policy. SMOPD addresses this tension by specializing teachers under complementary reward-priority profiles, then merging them through token-level on-policy distillation with a parameter-free uniform teacher mixture and a sequence-level anchor.
    \item \textbf{SMOPD improves both complementary and conflicting reward settings.}
    Across three backbones (1.5B, 3B, and 7B), SMOPD consistently exceeds GDPO across different model families. Specifically, in the complementary setting, the improvement is particularly pronounced—peaking at the 1.5B model with a \textbf{+48\%} composite gain and a dramatic format compliance jump from 8.8\% to \textbf{97.5\%}. In the conflicting setting, SMOPD achieves superior Safe Domain benchmark performance over GDPO across every setting.
    \item \textbf{The merged student surpasses its own teachers.} On safe alignment it exceeds both teachers and the scalarized baseline across three backbones; even at 7B, merging weak teachers yields a clear gain from complementary knowledge unused by scalarized training.
\end{itemize}

 \section{Related Work}

\subsection{Reinforcement Learning for LLMs}

RLHF aligns LLMs with human preferences \cite{christiano2017rlhf,ziegler2019finetuning,stiennon2020summarize,ouyang2022instructgpt}, classically with PPO \cite{schulman2017ppo} or its preference-based shortcut DPO \cite{rafailov2023dpo}, though optimizing against learned reward models is prone to over-optimization \cite{gao2023overoptimization}.
GRPO \cite{shao2024deepseekmath} removes the critic by normalizing rewards within a rollout group, and later variants sharpen this estimator, e.g.\ DAPO's decoupled clipping and dynamic sampling \cite{yu2025dapo}, GSPO's sequence-level importance ratios \cite{zheng2025gspo}, and simpler REINFORCE-style baselines \cite{ahmadian2024back}.
For multiple rewards, the standard treatment scalarizes them into one objective and refines the aggregation: GDPO \cite{liu2026gdpo} normalizes each reward separately and supports priority weights, DVAO \cite{dvao2026} adapts weights to per-reward gradient magnitudes, and SAW \cite{he2026saw} reweights objectives by learning speed; other lines constrain or trade off objectives instead of summing them \cite{xu2024cgpo,dai2024safe,zhou2023modpo,wang2024arithmetic}, or merge separately-trained policies in weight space \cite{wortsman2022soups,rame2024rewarded,rame2024warp}.
All of these ultimately ask a single set of parameters to absorb every reward at once, freezing one trade-off at training time; SMOPD instead trains one teacher per reward and defers balancing to a distillation-based merge.

\subsection{On-Policy Distillation}

Knowledge distillation originally trains a student on a teacher's soft targets \cite{hinton2015distilling}, extended to autoregressive LMs by sequence-level KD on teacher-generated text \cite{kim2016seqkd}.
Because teacher-generated data mismatches what the student sees at inference, GKD \cite{agarwal2024gkd} distills on the student's \emph{own} rollouts---on-policy distillation---with the divergence choice studied by MiniLLM \cite{gu2024minillm}, DistiLLM \cite{ko2024distillm}, and $f$-divergence KD \cite{wen2023fdistill}, and rollout selection refined by PG-OPD \cite{zhao2026pgopd}; we follow this line and use forward KL on student rollouts.
Recent work scales OPD to \emph{multiple teachers}, fusing separately-trained domain-specialized policies (math, coding, instruction following) into one model more effectively than reward mixing, cascade RL, or parameter merging \cite{ma2026mopd}; MiMo \cite{mimo2026flash} scales this recipe to frontier post-training, and G-OPD \cite{yang2026gopd} merges domain-specialized policies back into a shared base.
Because each prompt belongs to a single domain, all of these route it to the one teacher that owns it.
Our problem is orthogonal: \emph{within a single domain}, several rewards act on the \emph{same} prompt at once, so no routing can separate them; SMOPD instead combines reward-specialized teachers at every token, turning on-policy distillation into a mechanism for multi-reward balancing.

 \section{Method: SMOPD Framework}

The core philosophy of SMOPD is to let each reward be learned where it is easiest, in a \emph{teacher} trained to prioritize it, and then to transfer the teachers' competence into a single student at the granularity where rewards actually act: \emph{individual tokens}. This section presents the two stages in turn.

\subsection{Stage 1: Reward-Specialized Teacher Training}
\label{sec:teacher_training}

Group-based methods sample $G$ rollouts $\{o_1, \ldots, o_G\}$ per prompt and score each rollout $o_i$ with $K$ reward dimensions $r_i = (r_i^{(1)}, \ldots, r_i^{(K)})$; we write $\mu^{(k)}, \sigma^{(k)}$ for the group mean and standard deviation of dimension $k$.
GRPO \cite{shao2024deepseekmath} normalizes the \emph{summed} reward within the group, $\hat{A}_i^{\text{GRPO}} = (\sum_k r_i^{(k)} - \mu_S)/(\sigma_S + \epsilon)$, where $\mu_S, \sigma_S$ are the group statistics of the sum and $\epsilon$ is a small stabilizing constant.
GDPO \cite{liu2026gdpo} instead (i) normalizes each dimension within the group, (ii) aggregates the resulting advantages under priority weights $\mathbf{w} = (w_1, \ldots, w_K)$ (equal by default), and (iii) applies a final \emph{batch}-wise whitening over all responses in the update:
\begin{align}
\hat{A}_i^{(k)}
  &= \frac{r_i^{(k)} - \mu^{(k)}}{\sigma^{(k)}},
  \label{eq:gdpo_perreward}\\
A_i^{\text{sum}}
  &= \sum_{k=1}^K w_k \cdot \hat{A}_i^{(k)},
  \label{eq:gdpo_sum}\\
\hat{A}_i^{\text{GDPO}}
  &= \frac{A_i^{\text{sum}} - \mu_{\text{batch}}(A^{\text{sum}})}
  {\sigma_{\text{batch}}(A^{\text{sum}}) + \epsilon}.
  \label{eq:gdpo}
\end{align}
Both estimators train the policy $\pi_\theta$ with the standard clipped surrogate objective on these advantages \cite{schulman2017ppo,shao2024deepseekmath}.

GDPO's priority weights thus let a single policy be steered toward a selected trade-off. SMOPD repurposes this control as \emph{reward-priority specialization}: a set of complementary priority profiles constructs multiple teachers, one for each reward dimension, before they are merged in Stage~2.
For a reward dimension $k$ we wish to specialize on, we set $w_k \gg w_j$ for $j \neq k$:
\begin{equation}
\hat{A}_{\mathrm{teacher},k} = w_k^{\text{high}} \cdot \hat{A}^{(k)} + \sum_{j \neq k} w_j^{\text{low}} \cdot \hat{A}^{(j)}
\end{equation}

For example, with $K=2$ (accuracy and format), we train a \textbf{format teacher} with $\mathbf{w}=(0.1,0.9)$, amplifying the format reward signal by $9\times$ relative to accuracy, and an \textbf{accuracy teacher} with $\mathbf{w}=(0.9,0.1)$, and vice versa.

Because per-reward normalization gives each $\hat{A}^{(k)}$ zero mean and unit variance, the $9\times$ weight ratio directly translates to a $9\times$ expected gradient contribution from the favored reward (the normalized advantages are placed on a common scale, so the weights alone set their relative influence).
This makes the model ``zoom in'' on its target dimension, reliably maximizing it; the non-target dimensions may degrade (as for the accuracy teacher) or, when the dimensions are not in conflict, be retained (as for the format teacher, Section~\ref{sec:teachers}).
Either way the teachers become \emph{complementary}, each contributing a distinct strength to the subsequent merge.

\subsection{Stage 2: Multi-Teacher Online Policy Distillation}
\label{sec:opd}

Given $M$ teacher teachers $\{\pi_1, \ldots, \pi_M\}$, SMOPD distills their token-level knowledge into a student $\pi_\theta$.
Unlike offline distillation, the student generates its own responses (on-policy), and teachers provide top-$\kappa$ log-probabilities on the student's sequences---following the on-policy distillation paradigm \cite{agarwal2024gkd}, which avoids the exposure bias of distilling on a fixed teacher-generated corpus.

\paragraph{Teacher mixture distribution.}
At each token position $t$ in a student-generated response, the teacher target is the $\alpha$-weighted mixture of the $M$ teachers' next-token distributions:
\begin{equation}\label{eq:mixture}
p_T^{\text{mix}}(v \mid s_t) \;\propto\; \sum_{m=1}^M \alpha_{m} \, p_{\pi_m}(v \mid s_t),
\end{equation}
where $v$ ranges over vocabulary tokens, $s_t$ is the student's generated prefix up to position $t$, and $\alpha_m$ are the mixture weights. SMOPD uses \textbf{uniform} weights $\alpha_m = 1/M$ (e.g.\ $[0.5, 0.5]$ for two teachers), the simplest and, as we show, strongest choice in our setting; input-dependent gating alternatives are explored in Supplementary material~\ref{app:gates}.

Operating over the full vocabulary for every teacher at every position is prohibitive, so we realize Eq.~\eqref{eq:mixture} with a \emph{component-wise top-$\kappa$} construction ($\kappa = 16$ by default, distinct from the reward-dimension count $K$; sensitivity is studied in Section~\ref{sec:opd_ablation}). Each teacher $m$ emits only its own top-$\kappa$ token set $\mathcal{T}_m^\kappa$ with log-probabilities $\log p_{\pi_m}$. We shift each by $\log\alpha_m$, pool the $M\!\times\!\kappa$ candidates, and keep the $\kappa$ largest as the mixture target:
\begin{equation}\label{eq:mixture_impl}
\tilde{p}_{T}^{\text{mix}}(\cdot \mid s_t) =
\operatorname*{top\text{-}\kappa}_{\substack{m=1,\ldots,M\\ v \in \mathcal{T}_m^\kappa}}
\big[\, \log\alpha_m + \log p_{\pi_m}(v \mid s_t) \,\big].
\end{equation}
This is a fast approximation to the exact union-and-renormalize mixture ($\log\sum_m \alpha_m p_m$): a token in several teachers' top-$\kappa$ sets may appear more than once among the candidates, and tokens outside every teacher's top-$\kappa$ set are dropped. With high-quality teachers whose top-$\kappa$ mass dominates, the two coincide closely. The result is a single per-position top-$\kappa$ target that jointly carries both teachers.

\paragraph{Distillation loss.}
The student minimizes the forward KL divergence from the (top-$\kappa$) teacher mixture, evaluated over the mixture's support:
\begin{equation}\label{eq:opd_loss}
\cL_{\text{OPD}} = \sum_t \operatorname{KL}\left(\tilde{p}_{T}^{\text{mix}}(\cdot \mid s_t) \,\middle\|\, p_\theta(\cdot \mid s_t)\right)
\end{equation}
We use forward (rather than reverse) KL deliberately: forward KL is mode-covering \cite{gu2024minillm,ko2024distillm}, encouraging the student to place mass on \emph{all} high-probability tokens of the teacher mixture---important here because different teachers may be confident about different tokens (e.g., format wrappers vs.\ function arguments) at the same position, and the union-top-$\kappa$ target in Eq.~\eqref{eq:mixture_impl} preserves both.
Section~\ref{sec:opd_ablation} ablates both choices, while Supplementary material~\ref{app:opd_ablation} details the sampled-token reverse-KL estimator.

\paragraph{GDPO anchor.}
To prevent the student from only imitating teacher distributions without task-level grounding, we add a GDPO loss with equal weights $\mathbf{w}_{\text{equal}} = (1/K, \ldots, 1/K)$:
\begin{equation}
\cL_{\text{anchor}} = \cL_{\text{GDPO}}(\pi_\theta; \mathbf{w}_{\text{equal}}),
\end{equation}
where $\cL_{\text{GDPO}}(\pi_\theta; \mathbf{w})$ denotes the clipped policy-gradient loss driven by the GDPO advantage of Eq.~\eqref{eq:gdpo} under weights $\mathbf{w}$.

\paragraph{Total objective.}
The student optimizes:
\begin{equation}\label{eq:total_loss}
\cL_{\text{total}} = \cL_{\text{anchor}} + \lambda \cdot \cL_{\text{OPD}}
\end{equation}
where $\lambda$ controls the distillation strength (default $\lambda = 1.0$).
The anchor provides sequence-level direction (``this response should be upweighted/downweighted''), while OPD provides token-level guidance (``at this position, the distribution should look like this mixture of teachers'').

\begin{table*}[t]
\centering
\setlength{\tabcolsep}{3.6pt}
\begin{tabular}{ll ccc | c cccc}
\toprule
& & \multicolumn{3}{c|}{\textbf{RLLA-4K test} (in-domain)} & \multicolumn{5}{c}{\textbf{Generalization} (held-out tool use)} \\
\cmidrule(lr){3-5}\cmidrule(lr){6-10}
& & \textbf{Acc} & \textbf{Format} & \textbf{RLLA} & \textbf{BFCL} & \multicolumn{4}{c}{\textbf{API-Bank (LLM-Judge)} (\%)} \\
\cmidrule(lr){7-10}
\textbf{Backbone} & \textbf{Method} & \textbf{Reward} & \textbf{Pass} & \textbf{Mean} & \textbf{AST} & \textbf{L1} & \textbf{L2} & \textbf{L3} & \textbf{Avg} \\
\midrule
\multirow{6}{*}{Qwen2.5-1.5B} & GRPO Baseline & 1.737 & 5.0\% & 1.787 & 68.6\% & 79.2 & 71.6 & \underline{84.0} & 79.4 \\
 & GDPO Baseline & 1.761 & \underline{8.8\%} & 1.849 & \underline{72.6\%} & \underline{84.7} & \underline{74.6} & \textbf{84.7} & \underline{83.6} \\
 & GD$^2$PO Baseline & \textbf{1.776} & \underline{8.8\%} & \underline{1.864} & \textbf{73.3\%} & 82.2 & 71.6 & 82.4 & 81.1 \\
\cmidrule(l){2-10}
 & Accuracy Teacher [0.9,0.1] & 1.779 & 1.2\% & 1.791 & 70.5\% & 81.2 & 73.1 & 84.0 & 80.9 \\
 & Format Teacher [0.1,0.9] & 1.771 & 97.5\% & 2.746 & 70.5\% & 85.5 & 73.1 & 67.9 & 80.2 \\
\cmidrule(l){2-10}
\rowcolor{smopdrow}
 & SMOPD & \underline{1.765} & \textbf{97.5\%} & \textbf{2.740} & 70.7\% & \textbf{90.0} & \textbf{82.1} & 80.9 & \textbf{87.1} \\
\midrule
\multirow{6}{*}{Qwen2.5-3B} & GRPO Baseline & \textbf{1.783} & \textbf{97.5\%} & \textbf{2.758} & 75.8\% & 81.5 & 65.7 & 47.3 & 72.2 \\
 & GDPO Baseline & 1.753 & \textbf{97.5\%} & 2.728 & 75.3\% & 80.7 & \textbf{74.6} & \underline{64.1} & \underline{76.4} \\
 & GD$^2$PO Baseline & 1.732 & \textbf{97.5\%} & 2.707 & \underline{76.5\%} & \underline{82.0} & \underline{67.2} & 61.1 & 75.7 \\
\cmidrule(l){2-10}
 & Accuracy Teacher [0.9,0.1] & 1.795 & 95.0\% & 2.745 & 77.2\% & 83.2 & 68.7 & 61.8 & 76.9 \\
 & Format Teacher [0.1,0.9] & 1.740 & 97.2\% & 2.712 & 76.1\% & 84.2 & 70.2 & 62.6 & 77.9 \\
\cmidrule(l){2-10}
\rowcolor{smopdrow}
 & SMOPD & \underline{1.763} & \textbf{97.5\%} & \underline{2.738} & \textbf{77.2\%} & \textbf{83.5} & 64.2 & \textbf{65.7} & \textbf{77.4} \\
\bottomrule
\end{tabular}
\caption{Main results on complementary rewards setting (RLLA), across Qwen2.5-\{1.5B, 3B\}-Instruct. \textbf{In-domain} columns use the held-out RLLA-4K test split; \textbf{Generalization} columns are held-out tool-use benchmarks. For API-Bank, \mbox{\texttt{qwen3.7-plus}} serves as the semantic judge for original exact-match failures while exact-match successes are retained; All metrics are defined in Section~\ref{sec:setup-metrics}; \textbf{Bold} marks the best and \underline{underline} the second best among the baselines and SMOPD;}
\label{tab:acc_format}
\end{table*}
\begin{table*}[t]
\centering
\setlength{\tabcolsep}{4.5pt}
\small
\begin{tabular}{llccccccccccc}
\toprule
& & \multicolumn{3}{c}{\textbf{HH-RLHF}} & \multicolumn{3}{c}{\textbf{PKU-SafeRLHF}} & \multicolumn{3}{c}{\textbf{Alpaca}} & \textbf{Overall} \\
\cmidrule(lr){3-5} \cmidrule(lr){6-8} \cmidrule(lr){9-11} \cmidrule(lr){12-12}
\textbf{Backbone} & \textbf{Method} & U. & H. & Avg & U. & H. & Avg & U. & H. & Avg & \textbf{Avg} \\
\midrule
 \multirow{6}{*}{Qwen2.5-3B} & GRPO & 4.400 & 5.690 & 5.045 & 5.372 & 6.878 & 6.125 & 5.281 & 6.020 & 5.650 & 5.607 \\
 & GDPO & 4.382 & 5.704 & 5.043 & 5.373 & \textbf{6.883} & 6.128 & 5.247 & 6.001 & 5.624 & 5.598 \\
 & GD$^2$PO & \underline{4.485} & \underline{5.730} & \underline{5.107} & \underline{5.406} & \underline{6.879} & \underline{6.143} & \underline{5.379} & \underline{6.023} & \underline{5.701} & \underline{5.650} \\
\cmidrule(l){2-12}
 & Useful Teacher & 4.448 & 5.472 & 4.960 & 5.303 & 6.700 & 6.001 & 5.407 & 5.736 & 5.572 & 5.511 \\
 & Harmless Teacher & 4.320 & 5.784 & 5.052 & 5.375 & 6.919 & 6.147 & 5.149 & 6.176 & 5.662 & 5.620 \\
\cmidrule(l){2-12}
\rowcolor{smopdrow}
 & SMOPD & \textbf{4.511} & \textbf{5.743} & \textbf{5.127} & \textbf{5.426} & 6.869 & \textbf{6.147} & \textbf{5.392} & \textbf{6.071} & \textbf{5.732} & \textbf{5.669} \\
\midrule
 \multirow{6}{*}{Qwen2.5-7B} & GRPO & 4.237 & 5.469 & 4.853 & 5.386 & \underline{6.834} & \underline{6.110} & 4.976 & 5.647 & 5.312 & 5.425 \\
 & GDPO & 4.318 & 5.439 & 4.878 & 5.329 & 6.766 & 6.048 & 5.323 & 5.813 & 5.568 & 5.498 \\
 & GD$^2$PO & \underline{4.379} & \underline{5.511} & \underline{4.945} & \underline{5.390} & 6.801 & 6.096 & \textbf{5.434} & \underline{5.909} & \underline{5.672} & \underline{5.571} \\
\cmidrule(l){2-12}
 & Useful Teacher & 4.515 & 5.313 & 4.914 & 5.430 & 6.658 & 6.044 & 5.513 & 5.572 & 5.543 & 5.500 \\
 & Harmless Teacher & 4.220 & 5.607 & 4.913 & 5.354 & 6.889 & 6.122 & 5.021 & 6.041 & 5.531 & 5.522 \\
\cmidrule(l){2-12}
\rowcolor{smopdrow}
 & SMOPD & \textbf{4.475} & \textbf{5.656} & \textbf{5.065} & \textbf{5.448} & \textbf{6.837} & \textbf{6.143} & \underline{5.421} & \textbf{6.036} & \textbf{5.729} & \textbf{5.646} \\
\midrule
 \multirow{6}{*}{Llama-3.2-3B} & GRPO & 4.264 & 5.510 & 4.887 & 5.117 & 6.645 & 5.881 & 5.340 & 5.891 & 5.615 & 5.461 \\
 & GDPO & 4.414 & \underline{5.642} & 5.028 & 5.315 & \underline{6.761} & 6.038 & 5.366 & \underline{5.998} & 5.682 & 5.583 \\
 & GD$^2$PO & \textbf{4.438} & \textbf{5.657} & \textbf{5.047} & \underline{5.319} & \textbf{6.776} & \underline{6.047} & \textbf{5.429} & \textbf{6.012} & \textbf{5.721} & \textbf{5.605} \\
\cmidrule(l){2-12}
 & Useful Teacher & 4.174 & 5.177 & 4.676 & 5.198 & 6.512 & 5.855 & 5.296 & 5.489 & 5.393 & 5.308 \\
 & Harmless Teacher & 3.917 & 5.271 & 4.594 & 5.046 & 6.626 & 5.836 & 4.869 & 5.724 & 5.296 & 5.242 \\
\cmidrule(l){2-12}
\rowcolor{smopdrow}
 & SMOPD & \underline{4.437} & 5.633 & \underline{5.035} & \textbf{5.349} & 6.756 & \textbf{6.052} & \underline{5.406} & 5.962 & \underline{5.684} & \underline{5.590} \\
\bottomrule
\end{tabular}
\caption{Main results on conflicting rewards(Safe-alignment benchmarks) across Qwen2.5-{3B, 7B}-Instruct, and Llama-3.2-3B-Instruct, reported per benchmark with separate Useful (U.) and Harmless (H.) scores and their average (Avg); \textbf{Overall} is the mean Avg across the three benchmarks.}
\label{tab:safe_align}
\end{table*}

\section{Experiments}

\subsection{Experimental Setup}
\label{sec:setup-metrics}

\paragraph{Training.}
We train SMOPD on five backbones.
Within each setting, the baselines, reward-specialized teachers, and SMOPD students share the same training recipe, differing only in the GDPO priority profile and, for SMOPD, the added distillation loss; all runs utilize the verl framework \cite{sheng2024hybridflow} on 8 GPUs.
\textbf{Complementary rewards(RLLA).}
We train Qwen2.5-\{1.5B, 3B\}-Instruct \cite{qwen2024qwen25} with $G{=}8$ rollouts per prompt on the RLLA-4K \emph{training split} from ToolRL \cite{qian2025toolrl}.
The \emph{accuracy reward} ($r_{\text{acc}} \in [-3, 3]$) scores function-name and parameter matching, while the binary \emph{format reward} ($r_{\text{fmt}} \in \{0, 1\}$) checks the required XML structure (\texttt{<think>}, \texttt{<tool\_call>}, and \texttt{<response>} tags).
The format and accuracy teachers are trained with complementary priority profiles, $\mathbf{w}=(0.1,0.9)$ and $(0.9,0.1)$, respectively; we additionally conduct an ablation study on other priority configurations in Supplementary material~\ref{app:moderate}.
\textbf{Conflicting rewards(helpful + harmless).}
We train Qwen2.5-\{3B, 7B\}-Instruct and Llama-3.2-3B-Instruct with $G{=}4$ on prompt-only Alpaca \cite{taori2023alpaca}.
Useful and harmless rewards are produced by dual reward models trained on PKU-SafeRLHF preference data \cite{dai2024safe}, and the corresponding teachers employee the complementary profiles, $[0.7,0.3]$ and $[0.3,0.7]$.
Full training hyperparameters are provided in Supplementary material~\ref{app:details} (Tables~\ref{tab:hyperparams} and~\ref{tab:hyperparams_safe}).

\paragraph{Evaluation.}
All models are evaluated with vLLM \cite{kwon2023vllm} using temperature 0.0 and top-$p$ 1.0;
\textbf{Complementary rewards(RLLA).}
On the held-out RLLA-4K test split (80 prompts), \textbf{RLLA Mean} is the primary composite metric $r_{\text{acc}}+r_{\text{fmt}}$; we additionally report its \textbf{Acc Reward} component and the binary \textbf{Format Pass} rate.
Generalization is evaluated on two held-out tool-use benchmarks: \textbf{BFCL AST} is BFCL-v4 \cite{patil2023gorilla} function-calling accuracy averaged over the non-live and live AST categories, and \textbf{API-Bank (LLM-Judge)} \cite{li2023apibank} measures functional accuracy on Level-1/2/3.
For API-Bank, exact-match successes remain correct and an LLM judge \cite{zheng2023judging} reviews only exact-match failures for functionally equivalent tool calls; we report \textbf{L1}--\textbf{L3} and micro-accuracy over all 597 items (\textbf{Avg}).
The BFCL category breakdown is in Supplementary material~\ref{app:bfcl}, and the complete API-Bank protocol, original strict scores, and failure analysis are in Supplementary material~\ref{app:apibank}.
\textbf{Conflicting rewards(helpful + harmless).}
Every model is scored by the same dual reward models on held-out HH-RLHF \cite{bai2022hh}, PKU-SafeRLHF, and Alpaca prompts (mean@1).
We report per-benchmark \textbf{Useful}, \textbf{Harmless}, and their average; \textbf{Overall} is the mean of these averages across the three benchmarks.

\subsection{Main Results}
\paragraph{Reward-priority specialization makes sparse reward distributions learnable.}
\label{sec:teachers}
As shown in the 1.5B setting of Table~\ref{tab:acc_format} (rows 2--4), GDPO struggles to optimize the format reward under balanced priorities, achieving only 8.8\% format compliance on the RLLA test set.
This is because the binary format reward follows a sparse reward distribution: in most rollout groups, responses receive identical format scores, leaving little within-group variation for group-based RL to exploit.
Although the non-zero format score indicates that the model can occasionally generate correctly formatted responses, this sparse signal is overwhelmed by the denser accuracy reward distribution during balanced multi-reward optimization.
By increasing the priority of the format reward, the format-specialized teacher ($\mathbf{w}=[0.1,0.9]$) amplifies this otherwise underutilized signal and raises format compliance to 97.5\%.
Importantly, this gain does not come at the cost of accuracy performance. It shows that reward-priority specialization can make the sparse reward distribution learnable while preserving the dense-reward capability.

\paragraph{SMOPD under complementary rewards.}
We first study the complementary reward setting, where accuracy and format rewards supervise different aspects of tool-calling behavior.
Table~\ref{tab:acc_format} reports results on both backbones.
On Qwen2.5-1.5B, SMOPD achieves an RLLA score of 2.740, yielding a \textbf{+48\%} improvement over the GDPO baseline (1.849), while preserving the structured-output capability learned by the format-specialized teacher (97.5\% format compliance).
Meanwhile, the student retains accuracy-oriented tool-use ability: BFCL AST reaches 70.7\%, and API-Bank LLM-judge Avg reaches 87.1\%---the best among all 1.5B methods in Table~\ref{tab:acc_format} and \textbf{+3.5\%} points above the strongest scalarized baseline.
We report the API-Bank LLM-judge metric because the original strict exact-match protocol can penalize functionally equivalent tool calls due to superficial differences in argument formatting or values; therefore, we apply semantic judging only to exact-match failures while preserving exact successes.
The complete judging procedure and the corresponding strict exact-match results are provided in Supplementary material~\ref{app:apibank}.
These results show that SMOPD can merge complementary teacher capabilities instead of collapsing them into a single averaged behavior.

\paragraph{SMOPD under conflicting rewards.}
We further evaluate SMOPD under conflicting rewards, where helpfulness and harmlessness require balancing competing alignment objectives (Section~\ref{sec:teachers}).
Across three backbones (Table~\ref{tab:safe_align}), SMOPD consistently improves over scalarized baselines by merging reward-specialized capabilities; adaptive teacher mixing variants are compared in Supplementary material~\ref{app:gates}.
On Qwen2.5-3B, SMOPD achieves the best Overall score (\textbf{5.669}), outperforming the strongest scalarized baseline GD$^2$PO (5.650) and both single-reward teachers.
The gain becomes more evident on Qwen2.5-7B: although each individual teacher provides only marginal improvement over GDPO (5.500 and 5.522 versus 5.498), SMOPD further improves the merged policy to 5.646, exceeding all baselines and teachers.
This suggests that even under conflicting objectives, reward-specialized teachers retain complementary capabilities that can be recovered through policy-level merging.
On Llama-3.2-3B, where teacher specialization is the most challenging, both individual teachers degrade substantially, yet SMOPD recovers the performance to 5.590, approaching the strongest scalarized baseline GD$^2$PO (5.605) while surpassing GDPO (5.583).
The remaining gap to GD$^2$PO is mainly attributed to the task anchor rather than the teacher-merging mechanism: SMOPD uses GDPO as its sequence-level anchor, whereas GD$^2$PO improves the scalarized optimization objective itself through a refined advantage estimator.
Therefore, improving the anchor provides an orthogonal direction to further enhance SMOPD, while the consistent gain over GDPO demonstrates the effectiveness of reward-specialized capability merging.

 \subsection{Ablation and Analysis}

\paragraph{Ablating the Task Anchor.}
\label{sec:two_layers}

SMOPD combines two training signals operating at different levels of granularity.
The \textit{sequence-level} GDPO anchor determines whether a sampled response should be upweighted or downweighted according to reward-normalized advantages, providing task-level optimization direction but only a coarse scalar signal for the entire response.
In contrast, \textit{token-level} OPD distillation shapes the student's distribution toward the teacher mixture at every generation step, providing dense supervision through hundreds of token-level gradient signals within a single response.

This combination is particularly valuable for format compliance, where the reward is binary (0/1 per response) but the required behavior---correct XML structure spanning many tokens---needs position-level guidance.
The GDPO anchor alone cannot teach format because it reduces to a single advantage shared by all tokens in a response.
OPD provides the missing token-level granularity: the format teacher's high confidence on XML wrapper tokens (\texttt{<tool\_call>}, \texttt{</tool\_call>}) and the accuracy teacher's specialization on function-argument tokens jointly shape the student's behavior at each position.

To test whether both signals are necessary, we ablate the anchor on Qwen2.5-7B safe alignment at the controlled setting $\kappa=32$, optimizing only the OPD objective, $\cL_{\text{total}}=\cL_{\text{OPD}}$ (Table~\ref{tab:anchor_ablation}).
OPD alone already improves over the individual teachers, reaching $5.544$ Overall compared with $5.500$ for the useful teacher and $5.522$ for the harmless teacher.
However, it remains close to teacher performance, suggesting that pure distillation is limited by the quality of the teacher mixture it imitates.
Reintroducing the anchor lifts SMOPD to $5.639$ Overall, exceeding both teachers.
The anchor therefore supplies an additional task-level optimization signal beyond teacher imitation, allowing the student to improve after absorbing complementary teacher behaviors.
Supplementary material~\ref{app:anchor_theory} formalizes this distinction: the forward-KL objective in Eq.~\eqref{eq:opd_loss} reaches its minimum at the teacher mixture, where the distillation gradient vanishes, whereas the policy-gradient anchor can continue optimizing task reward.

\begin{table}[h]
\centering
\small
\setlength{\tabcolsep}{5pt}
\begin{tabular}{lccc}
\toprule
\textbf{Method} & \textbf{Useful} & \textbf{Harmless} & \textbf{Overall} \\
\midrule
GDPO baseline (anchor only) & 4.990 & 6.006 & 5.498 \\
Useful teacher & 5.153 & 5.848 & 5.500 \\
Harmless teacher & 4.865 & 6.179 & 5.522 \\
\midrule
OPD only (no anchor) & \underline{5.011} & \underline{6.077} & \underline{5.544} \\
\rowcolor{defaultrow}
SMOPD (OPD $+$ anchor) & \textbf{5.099} & \textbf{6.178} & \textbf{5.639} \\
\bottomrule
\end{tabular}
\caption{Ablating the task anchor on Qwen2.5-7B-Instruct safe alignment at the controlled setting $\kappa=32$ (dual-RM mean@1; \textbf{Overall} $=$ mean of Useful and Harmless).}
\label{tab:anchor_ablation}
\end{table}

\paragraph{Ablating OPD Design Choices.}
\label{sec:opd_ablation}

\begin{figure}[!t]
\centering
\includegraphics[width=\columnwidth]{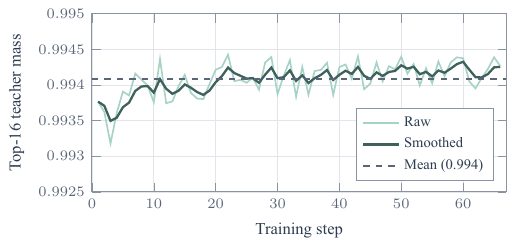}
\caption{Top-16 teacher probability mass during Qwen2.5-7B safe-alignment training; light and dark curves show the raw and smoothed values.}
\label{fig:topk16_mass}
\end{figure}

The top-$\kappa$ approximation controls the number of teacher candidates retained at each position, trading a broader approximation to the teacher distributions for distillation cost.
We vary $\kappa\in\{16,32,64\}$ for SMOPD on Qwen2.5-7B-Instruct safe alignment while keeping the training and evaluation protocol unchanged, and separately compare the sampled-token reverse-KL update detailed in Supplementary material~\ref{app:opd_ablation} (Table~\ref{tab:opd_ablation}).
Forward-KL performance remains highly stable across support sizes, with Overall varying by only $0.007$ between $\kappa=16$ and $\kappa=64$.
To understand why a small support is sufficient, we measure the teacher probability mass retained by the top-$\kappa$ approximation. The top-16 support already preserves approximately $0.994$ of the teacher mass throughout training (Figure~\ref{fig:topk16_mass}), explaining why increasing $\kappa$ provides little additional benefit.

The sampled-token reverse-KL variant reaches 5.563 Overall, $0.082$ below forward KL with $\kappa=16$.
This gap is consistent with the different behaviors of the two divergence directions in our multi-teacher setting.
Forward KL is mode-covering, encouraging the student to preserve the diverse high-probability regions in the teacher mixture, which is important when different teachers contribute distinct behaviors.
In contrast, reverse KL is mode-seeking and may concentrate on a subset of dominant teacher modes, potentially losing specialized capabilities from other teachers.
Thus, forward KL is preferable in this setting, although this single-backbone result does not imply that reverse KL is universally inferior.

\begingroup
\begin{table}[h]
\centering
\small
\setlength{\tabcolsep}{3.5pt}
\begin{tabular}{llccc}
\toprule
\textbf{Objective} & \textbf{Support} & \textbf{Useful} & \textbf{Harmless} & \textbf{Overall} \\
\midrule
GDPO baseline & -- & 4.990 & 6.006 & 5.498 \\
\midrule
\rowcolor{defaultrow}
Forward KL & $\kappa=16$ & \textbf{5.115} & 6.176 & \textbf{5.646} \\
Forward KL & $\kappa=32$ & 5.099 & \underline{6.178} & 5.639 \\
Forward KL & $\kappa=64$ & \underline{5.102} & \textbf{6.181} & \underline{5.642} \\
\midrule
Reverse KL (PG) & sampled & 5.034 & 6.093 & 5.563 \\
\bottomrule
\end{tabular}
\caption{Top-$\kappa$ support and KL-direction ablation on Qwen2.5-7B safe alignment. Useful and Harmless are averaged across HH-RLHF, PKU-SafeRLHF, and Alpaca; Overall is their mean.}
\label{tab:opd_ablation}
\end{table}
\endgroup

\section{Conclusion}
We identified a reward-density imbalance in multi-reward reinforcement learning: even after per-reward normalization, sparse rewards may provide too little within-group variation to effectively influence a shared policy.
To overcome this limitation, we introduced SMOPD, which decouples capability acquisition from reward balancing by training reward-specialized teachers and merging their capabilities into a single student through online policy distillation with a balanced task anchor.
Across complementary rewards and conflicting rewards settings, SMOPD consistently improves over GDPO on 1.5B, 3B, and 7B backbones.
% In particular, SMOPD raises format compliance from $8.8\%$ to $97.5\%$ and improves the composite RLLA score by $48\%$ at 1.5B, while also improving safe-alignment performance across all evaluated backbones.
These results highlight the effectiveness of our method in multi-reward optimization: reward-specific specialization preserves complementary capabilities, while subsequent merging produces a balanced final policy without over-prioritizing any single objective. 
% The consistent gains across heterogeneous reward settings empirically validate our design.

\bibliographystyle{conference}
\bibliography{mopd}

\appendix
\clearpage

\section{Training and Evaluation Details}
\label{app:details}

All models are trained with the verl framework using Ray-based distributed training. Table~\ref{tab:hyperparams} lists the configuration of the accuracy+format (RLLA) experiments and Table~\ref{tab:hyperparams_safe} that of the safe-alignment experiments. Within each setting, GDPO, the teachers, and the SMOPD students share the same recipe, differing only in the GDPO reward weights (and, for SMOPD, the added distillation loss); GD$^2$PO changes only the advantage estimator, as detailed below.
The RLLA runs use 150 steps for Qwen2.5-1.5B and 100 for Qwen2.5-3B; safe-alignment runs use 100 steps for the 3B backbones and 66 steps for Qwen2.5-7B.
The confidence-failure gates (Sec.~\ref{app:gates}) use weight floor $\phi=0.2$, exponents $\gamma=\eta=1$, confidence floor $c_0=0.05$, blend strength $\beta=0.5$, and $\epsilon=10^{-6}$ in all settings.

\paragraph{GD$^2$PO baseline.}
The GD$^2$PO rows in Tables~\ref{tab:acc_format} and~\ref{tab:safe_align} use the hard conflict-filtering variant of GD$^2$PO~\citep{liu2026gd2po}.
%, implemented as \texttt{gdpo\_filter\_group\_ratio}.
Starting from GDPO's separately group-normalized advantage for each reward, GD$^2$PO-Hard removes a rollout whenever its nonzero reward-wise scalar advantages contain opposing signs. It then scales each prompt group's surviving advantages by that group's retained-rollout fraction and whitens over retained response tokens only. Reward weights are equal, and every other model, data, rollout, and optimization setting is identical to the corresponding GDPO run. %We do not use the alternative SNR-thresholded filter. 

\begin{table*}[t]
\begin{minipage}[t]{0.48\textwidth}
\centering
\small
\setlength{\tabcolsep}{3.5pt}
\begin{tabular}{lc}
\toprule
\textbf{Accuracy + format (RLLA)} & \textbf{Value} \\
\midrule
Base models & Qwen2.5-\{1.5B, 3B\}-Instruct \\
Training data & RLLA-4K train split (ToolRL) \\
Reward dimensions & $r_{\text{acc}} \in [-3,3]$, $r_{\text{fmt}} \in \{0,1\}$ \\
GD$^2$PO variant & hard sign-conflict filtering \\
Teacher weights & $[0.9,0.1]$ acc / $[0.1,0.9]$ fmt \\
Anchor weights & equal, $[0.5, 0.5]$ \\
Teacher mixing (fixed) & $\alpha_m = 1/M = 0.5$ \\
Training steps & 150 (1.5B); 100 (3B) \\
Rollouts per prompt ($G$) & 8 \\
Learning rate & $1 \times 10^{-6}$ \\
OPD top-$\kappa$ & 16 \\
OPD coefficient ($\lambda$) & 1.0 \\
Distillation loss & forward KL, top-$\kappa$ support \\
Eval temperature / top-$p$ & 0.0 / 1.0 \\
Max new tokens & 1024 \\
Eval protocol & greedy decoding, single pass \\
GPUs / tensor parallelism & 8 / 1 \\
Hardware & 8 GPUs \\
\bottomrule
\end{tabular}
\caption{Training and evaluation configuration for the accuracy + format experiments (both backbones).}
\label{tab:hyperparams}
\end{minipage}
\hfill
\begin{minipage}[t]{0.48\textwidth}
\centering
\small
\setlength{\tabcolsep}{2pt}
\begin{tabular}{@{}p{0.42\linewidth}p{0.54\linewidth}@{}}
\toprule
\textbf{Safe alignment (helpful + harmless)} & \textbf{Value} \\
\midrule
Base models & Qwen2.5-\{3B, 7B\}-Instruct, \\
& Llama-3.2-3B-Instruct \\
Training prompts & Alpaca (prompt-only) train split \\
Useful reward & Qwen2.5-7B-SafeRLHF-RM \\
Harmless reward & Qwen2.5-7B-SafeRLHF-CM \\
GD$^2$PO variant & hard sign-conflict filtering \\
Teacher weights & $[0.7,0.3]$ useful / \\
& $[0.3,0.7]$ harmless \\
Anchor weights & equal, $[0.5, 0.5]$ \\
Teacher mixing (fixed) & $\alpha_m = 1/M = 0.5$ \\
Train batch / mini-batch & 512 / 128 \\
Rollouts per prompt & 4 \\
Training steps & 100 (3B); 66 (7B) \\
Rollout temperature / top-$p$ & 0.7 / 1.0 \\
Learning rate & $1 \times 10^{-6}$ \\
OPD top-$\kappa$, $\lambda$ & 16, 1.0 \\
Eval benchmarks & HH-RLHF (8{,}520), \\
& PKU-SafeRLHF (8{,}211), Alpaca (512) \\
Eval protocol & mean@1 per benchmark \\
Hardware & 8 GPUs \\
\bottomrule
\end{tabular}
\caption{Training and evaluation configuration for the safe-alignment experiments (all three backbones).}
\label{tab:hyperparams_safe}
\end{minipage}
\end{table*}

% \paragraph{Early stopping on 7B safe alignment.}
% All 3B safe-alignment models train for 100 steps. On Qwen2.5-7B, however, training past roughly two-thirds of that schedule shows a clear \emph{reward-hacking} trend---the dual reward models keep rising while response quality degrades---so we set the training budget to 66 steps for every method. This shared 66-step schedule applies to the baselines, reward-specialized teachers, and SMOPD students. All 7B numbers in Table~\ref{tab:safe_align} therefore use the same training budget, so the merging gains are not an artifact of running SMOPD for a different number of steps than its baselines.

 \section{Why the Anchor Breaks the Teacher Ceiling}
\label{app:anchor_theory}

This supplementary material formalizes the claim of Section~\ref{sec:two_layers} that the two training signals of Eq.~\eqref{eq:total_loss} play distinct, complementary roles.
The forward-KL OPD loss can at best reproduce the teacher mixture, whereas the anchor's policy gradient is the only term that remains active at that point.
Moreover, on a sparse reward, the anchor becomes informative only after OPD has lifted the student's success rate.
Throughout, we treat the frozen teacher mixture of Eq.~\eqref{eq:mixture} as a fixed conditional distribution $q(\cdot \mid s)$, write $d_\theta$ for the state (prefix) distribution induced by the student's own rollouts, and ignore PPO-style clipping, which is inactive at the on-policy point where the importance ratio equals one.

\paragraph{Pure OPD is capped at the teacher mixture.}
The on-policy distillation objective of Eq.~\eqref{eq:opd_loss} is
\begin{equation}\label{eq:app_opd}
\cL_{\text{OPD}}(\theta) = \E_{s \sim d_\theta}\!\left[\operatorname{KL}\!\big(q(\cdot \mid s) \,\big\|\, p_\theta(\cdot \mid s)\big)\right].
\end{equation}
Since $\operatorname{KL}(q \,\|\, p_\theta) \ge 0$ with equality iff $p_\theta(\cdot \mid s) = q(\cdot \mid s)$ on the support of $q$, any student matching the mixture on its own visited states attains the global minimum $\cL_{\text{OPD}} = 0$.
Moreover, this minimum is a stationary point: the pointwise gradient is
\begin{equation}\label{eq:app_opd_grad}
\nabla_\theta \operatorname{KL}\big(q \,\|\, p_\theta\big)
= -\sum_v q(v \mid s)\, \nabla_\theta \log p_\theta(v \mid s),
\end{equation}
which at $p_\theta = q$ equals $-\sum_v \nabla_\theta\, p_\theta(v \mid s) = -\nabla_\theta 1 = 0$; the contribution of $\nabla_\theta d_\theta$ vanishes as well because the integrand is pointwise zero at the minimum.
Gradient descent on $\cL_{\text{OPD}}$ alone therefore terminates at the mixture policy: the student inherits the teachers' behavior but receives \emph{no} signal that would push its expected task reward above that of the mixture.
This is the imitation ceiling observed in Table~\ref{tab:anchor_ablation}, where the OPD-only student reaches $5.544$ Overall, barely above the teachers it imitates ($5.500$/$5.522$).

\paragraph{The anchor's gradient survives at the ceiling.}
The anchor is a policy-gradient term driven by the GDPO advantage $\hat{A}^{\text{GDPO}}$ of Eq.~\eqref{eq:gdpo}:
\begin{equation}\label{eq:app_anchor_grad}
-\nabla_\theta \cL_{\text{anchor}}
= \E_{o \sim \pi_\theta}\!\left[\hat{A}^{\text{GDPO}}(o)\, \nabla_\theta \log \pi_\theta(o)\right],
\end{equation}
an ascent direction on the group-normalized task reward.
Its stationary points are local optima of the reward, not matches to any teacher; in particular, the mixture $q$ is not a stationary point of $\cL_{\text{anchor}}$ unless $q$ already locally maximizes the task reward.
Hence at the OPD fixed point $p_\theta = q$,
\begin{equation}
\nabla_\theta \cL_{\text{total}}
= \nabla_\theta \cL_{\text{anchor}} + \lambda \cdot 0
= \nabla_\theta \cL_{\text{anchor}} \neq 0
\end{equation}
in general, so optimization of Eq.~\eqref{eq:total_loss} continues \emph{through} the imitation ceiling in the direction that increases task reward---exactly the $+0.095$ Overall lift of SMOPD over its anchor-free counterpart in Table~\ref{tab:anchor_ablation}.

\paragraph{Under a sparse reward, OPD is what activates the anchor.}
Why not rely on the anchor alone, then?
For a binary reward with per-rollout success probability $p^{\text{succ}}_\theta$ under the current policy, a group of $G$ i.i.d.\ rollouts receives identical rewards on that dimension with probability $(p^{\text{succ}}_\theta)^G + (1 - p^{\text{succ}}_\theta)^G$, and by Eq.~\eqref{eq:gdpo_perreward} such a degenerate group contributes exactly zero advantage on that dimension.
The anchor's expected signal on the sparse dimension is therefore proportional to the probability of a \emph{mixed} group,
\begin{equation}
1 - (p^{\text{succ}}_\theta)^G - (1 - p^{\text{succ}}_\theta)^G \;\approx\; G\, p^{\text{succ}}_\theta
\quad (p^{\text{succ}}_\theta \to 0),
\end{equation}
which vanishes linearly with the success rate: near $p^{\text{succ}}_\theta \approx 0$ the anchor is inert on exactly the dimension that needs it most.
The OPD gradient of Eq.~\eqref{eq:app_opd_grad}, by contrast, is \emph{dense}: it is nonzero at every position where the student deviates from the mixture, independent of within-group reward variance, and the format teacher concentrates its mass precisely on the wrapper tokens the student is missing.
Distillation therefore raises $p^{\text{succ}}_\theta$ rapidly; once the success rate is bounded away from $0$ and $1$, mixed groups occur with constant probability, the sparse dimension's advantage becomes non-degenerate, and the anchor's ascent direction of Eq.~\eqref{eq:app_anchor_grad} takes over.

In summary, the two losses are complementary by construction: OPD supplies the dense, reward-independent gradient that carries the student to the teachers' level and activates the sparse dimension's group signal, while the anchor is the only term whose gradient survives at the imitation ceiling---and is thus what pushes the student \emph{beyond} its teachers.

\section{Sampled-Token Reverse-KL Estimator}
\label{app:opd_ablation}

Section~\ref{sec:two_layers} compares forward KL against reverse KL while keeping uniform teacher mixing, the GDPO anchor, the training recipe, and the evaluation checkpoint fixed.
Because the reverse direction places the student in the first argument, we implement it in its natural on-policy, sampled-token form \citep{gu2024minillm,lu2025onpolicy}.
For each rollout token $y_t\sim\pi_{\mathrm{old}}$, every teacher returns the scalar log-probability of that token and the mixture is formed exactly:
\begin{equation}
\log q_{\mathrm{mix}}(y_t)
= \operatorname{logsumexp}_{m}\!\left(\log\alpha_m + \log p_{\pi_m}(y_t)\right),
\label{eq:revkl_mix}
\end{equation}
where conditioning on the prefix $s_t$ is implicit.
Let $D_{\mathrm{rev}}(\theta):=D_{\mathrm{KL}}(\pi_\theta\Vert q_{\mathrm{mix}})$.
The desired reverse divergence and its score-function gradient \citep{williams1992reinforce} are
\begin{align}
D_{\mathrm{rev}}(\theta)
&= \E_{y\sim\pi_\theta}\!\left[\log\pi_\theta(y)-\log q_{\mathrm{mix}}(y)\right], \nonumber\\
\nabla_\theta D_{\mathrm{rev}}(\theta)
&= \E_{y\sim\pi_\theta}\!\left[\delta_\theta(y)\nabla_\theta\log\pi_\theta(y)\right], \nonumber\\
\delta_\theta(y)
&= \log\pi_\theta(y)-\log q_{\mathrm{mix}}(y).
\label{eq:revkl_grad}
\end{align}
Accordingly, we detach the sampled log-ratio and use it as a token-level advantage in a clipped policy-ratio update:
\begin{align}
k_1(y_t) &= \log\pi_{\mathrm{old}}(y_t)-\log q_{\mathrm{mix}}(y_t), \nonumber\\
A_t &= -\operatorname{stopgrad}\!\left[k_1(y_t)\right], \nonumber\\
\rho_t &= \frac{\pi_\theta(y_t)}{\pi_{\mathrm{old}}(y_t)}, \nonumber\\
\cL_{\mathrm{rev\text{-}PG}}
&= -\E_t\!\left[\operatorname{clip}(\rho_t,1-\epsilon,1+\epsilon)A_t\right].
\label{eq:revkl_pg}
\end{align}
At the on-policy point $\pi_\theta=\pi_{\mathrm{old}}$, the expected gradient of Eq.~\eqref{eq:revkl_pg} matches Eq.~\eqref{eq:revkl_grad}.

This detached policy-gradient construction is essential.
Although a sampled $k_3$ scalar \citep{schulman2020kl} has the reverse-KL value in expectation, directly differentiating it while treating the rollout sampling distribution as fixed does not recover the reverse-KL gradient; it therefore cannot serve as a valid direction ablation.
We exclude that invalid control and report only the sampled-token estimator in Table~\ref{tab:opd_ablation}.
The sampled-token reverse-KL update remains below forward KL in this setting, but the comparison does not establish that reverse KL is universally inferior.

\section{Ablation: Sensitivity to the Priority Profile}
\label{app:moderate}

\begin{table*}[t]
\centering
\setlength{\tabcolsep}{4.5pt}
\begin{tabular}{ll ccc | c}
\toprule
& & \multicolumn{3}{c|}{\textbf{RLLA-4K test} (in-domain)} & \textbf{Gen.} \\
\cmidrule(lr){3-5}\cmidrule(lr){6-6}
\textbf{Backbone} & \textbf{Method} & \textbf{Acc} & \textbf{Format} & \textbf{RLLA} & \textbf{BFCL} \\
& & \textbf{Reward} & \textbf{Pass} & \textbf{Mean} & \textbf{AST} \\
\midrule
\multirow{4}{*}{Qwen2.5-3B}
 & GDPO Baseline (ref) & 1.753 & 97.5\% & 2.728 & 75.3\% \\
\cmidrule(l){2-6}
 & Accuracy Teacher [0.7,0.3] & 1.763 & 94.4\% & 2.707 & 76.4\% \\
 & Format Teacher [0.3,0.7] & 1.734 & 97.5\% & 2.709 & 73.4\% \\
\cmidrule(l){2-6}
 & SMOPD & \textbf{1.782} & \textbf{97.5\%} & \textbf{2.757} & \textbf{76.9\%} \\
\bottomrule
\end{tabular}
\caption{\emph{Moderate}-skew ablation on Qwen2.5-3B: teachers use softened GDPO priority profiles rather than the aggressive skew of the main results in Table~\ref{tab:acc_format}. API-Bank was not run for these moderate checkpoints and is therefore omitted. The GDPO baseline is reproduced for reference; \textbf{bold} marks the moderate-profile SMOPD result.}
\label{tab:acc_format_moderate}
\end{table*}

Our main experiments specialize teachers with aggressive priority profiles ($[0.9,0.1]/[0.1,0.9]$).
Here we ablate the \emph{strength} of that skew, re-running the entire acc+format pipeline on Qwen2.5-3B with softened \emph{moderate} weights ($[0.7,0.3]$ accuracy-heavy, $[0.3,0.7]$ format-heavy); Table~\ref{tab:acc_format_moderate} reports the results.
The moderate weights yield two competitive, complementary teachers ($94.4\%$/$97.5\%$ format), and merging works at least as well as under aggressive weights: moderate SMOPD attains $2.757$---the best result among all 3B acc+format runs---beating both of its own teachers ($2.707$/$2.709$), the GDPO baseline ($2.728$), and the aggressive-profile student ($2.738$).
The advantage of SMOPD is thus robust to how sharply the rewards are skewed: as long as the teachers remain complementary, merging surpasses the scalarized baseline, and the softer skew performs, if anything, marginally better.

 \section{Adaptive Teacher-Mixing Gates}
\label{app:gates}

SMOPD's default mixes teachers with \emph{uniform} weights $\alpha_m = 1/M$ (Section~\ref{sec:opd}). Here we ask whether a more adaptive, input-dependent gate could do better by routing more weight to the teacher that is most relevant at each sequence or token, and we define two \emph{confidence-failure} gates that steer the uniform prior $1/M$ using three signals derived from the teachers and rewards: each teacher's \emph{confidence} (how peaked its top-$\kappa$ distribution is), the student's \emph{reward failure} on that teacher's target dimension, and inter-teacher \emph{disagreement}.

\paragraph{Signals.}
At token $t$, teacher $m$'s confidence combines the normalized negative entropy and the top-two margin of its (top-$\kappa$, renormalized) distribution $p_{\pi_m}(\cdot\mid s_t)$:
\begin{equation}\label{eq:gate_conf}
c_{m,t} = \tfrac{1}{2}\Big(1 - \tfrac{H\!\left[p_{\pi_m}(\cdot\mid s_t)\right]}{\log\kappa}\Big) + \tfrac{1}{2}\big(p_{\pi_m}^{(1)}(s_t) - p_{\pi_m}^{(2)}(s_t)\big),
\end{equation}
where $H[\cdot]$ is Shannon entropy and $p^{(1)}\!\ge p^{(2)}$ are the two largest probabilities. Reward failure uses the response's reward $r_m$ on teacher $m$'s dimension, min--max normalized to $[0,1]$:
\begin{equation}\label{eq:gate_fail}
f_m = 1 - \operatorname{clamp}\!\Big(\tfrac{r_m - r_{\min}}{r_{\max}-r_{\min}},\,0,\,1\Big),
\end{equation}
where $[r_{\min}, r_{\max}]$ is the reward range of that dimension; a teacher is thus upweighted exactly where the student is failing its reward. Disagreement $d_t\in[0,1]$ is the fraction of teacher pairs whose top-1 tokens differ at $s_t$.

\paragraph{Sequence-level gate.}
Averaging $c_{m,t}$ over the response mask gives $\bar c_m$, and the raw weight combines the three signals over the uniform prior:
\begin{equation}\label{eq:gate_seq}
\tilde\alpha_m = \tfrac{1}{M}\,(\bar c_m + c_0)^{\eta}\,(f_m + \epsilon)^{\gamma},
\quad
\alpha_m = \frac{\tilde\alpha_m}{\sum_{j}\tilde\alpha_j}.
\end{equation}
Averaging disagreement to $\bar d$, we blend back toward the prior and apply a floor $\phi$ so no teacher is silenced:
\begin{equation}\label{eq:gate_blend}
\begin{aligned}
\alpha_m
&\leftarrow \Phi_\phi\!\left[(1-\beta\bar d)\,\alpha_m
   + \beta\bar d\cdot\tfrac{1}{M}\right],\\
\Phi_\phi[x]_m
&= \phi + (1-\phi M)\,\tfrac{x_m}{\sum_j x_j}.
\end{aligned}
\end{equation}

\paragraph{Token-level gate.}
The same construction is applied \emph{per position}: $c_{m,t}$ and $d_t$ are used directly (no averaging), yielding position-specific weights $\alpha_{m,t}$ that let different teachers dominate at different tokens. Reward failure $f_m$ remains sequence-level (one reward per response). Gate hyperparameters are fixed across all settings and listed in Sec.~\ref{app:details}.

\paragraph{Results.}
Tables~\ref{tab:gate_rlla} and~\ref{tab:gate_safe} compare the two gates against SMOPD's uniform mixing on all five settings. The 1.5B RLLA block uses the aggressive main profile, while the 3B RLLA block uses the moderate-profile ablation of Table~\ref{tab:acc_format_moderate}. Two findings stand out. First, \textbf{the three variants stay within a narrow band}: on RLLA Mean the spread never exceeds $0.06$, and on safety Overall it never exceeds $0.02$. Second, \textbf{uniform mixing is the strongest RLLA rule in both displayed profile settings}: it reaches 2.740 at 1.5B and 2.757 in the moderate 3B ablation, compared with 2.684/2.690 and 2.733/2.730 for the sequence/token gates. Token-level routing wins only on Qwen2.5-7B safe alignment (5.651 vs.\ 5.646), while uniform ties or leads in the remaining safety settings. The adaptive mechanisms therefore add no consistent benefit over the parameter-free mixture, motivating uniform mixing as SMOPD's default.

\begin{table*}[t]
\centering
\setlength{\tabcolsep}{7pt}
\begin{tabular}{ll ccc | c}
\toprule
& & \multicolumn{3}{c|}{\textbf{RLLA-4K test} (in-domain)} & \textbf{Generalization} \\
\cmidrule(lr){3-5}\cmidrule(lr){6-6}
& & \textbf{Acc} & \textbf{Format} & \textbf{RLLA} & \textbf{BFCL} \\
\textbf{Backbone} & \textbf{Mixing} & \textbf{Reward} & \textbf{Pass} & \textbf{Mean} & \textbf{AST} \\
\midrule
 \multirow{3}{*}{1.5B} & SMOPD (uniform) & \underline{1.765} & \textbf{97.5\%} & \textbf{2.740} & \underline{70.7\%} \\
 & SMOPD (seq gate) & 1.741 & \underline{94.3\%} & 2.684 & \textbf{70.9\%} \\
 & SMOPD (token gate) & \textbf{1.787} & 90.3\% & \underline{2.690} & \textbf{70.9\%} \\
\midrule
 \multirow{3}{*}{3B} & SMOPD (uniform) & \textbf{1.782} & \textbf{97.5\%} & \textbf{2.757} & \underline{76.9\%} \\
 & SMOPD (seq gate) & \underline{1.765} & \underline{96.9\%} & \underline{2.733} & 76.6\% \\
 & SMOPD (token gate) & 1.755 & \textbf{97.5\%} & 2.730 & \textbf{77.1\%} \\
\bottomrule
\end{tabular}
\caption{Teacher-mixing comparison on the accuracy + format setting. Rows are SMOPD with uniform mixing (our main method) and the two confidence-failure gates. The 1.5B rows use aggressive teachers; the 3B rows use the moderate teachers from the ablation in Table~\ref{tab:acc_format_moderate}, not the aggressive-profile 3B checkpoints in the main table. Per column and per backbone, \textbf{bold} marks the best and \underline{underline} the second best among the three mixing rules (ties share the rank).}
\label{tab:gate_rlla}
\end{table*}

\begin{table*}[t]
\centering
\setlength{\tabcolsep}{4.5pt}
\small
\begin{tabular}{llccccccccccc}
\toprule
& & \multicolumn{3}{c}{\textbf{HH-RLHF}} & \multicolumn{3}{c}{\textbf{PKU-SafeRLHF}} & \multicolumn{3}{c}{\textbf{Alpaca}} & \textbf{Overall} \\
\cmidrule(lr){3-5} \cmidrule(lr){6-8} \cmidrule(lr){9-11} \cmidrule(lr){12-12}
\textbf{Backbone} & \textbf{Mixing} & U. & H. & Avg & U. & H. & Avg & U. & H. & Avg & \textbf{Avg} \\
\midrule
 \multirow{3}{*}{Qwen2.5-3B} & SMOPD (uniform) & \textbf{4.511} & 5.743 & \underline{5.127} & \textbf{5.426} & 6.869 & \underline{6.147} & \textbf{5.392} & \underline{6.071} & \textbf{5.732} & \textbf{5.669} \\
 & SMOPD (seq gate) & \underline{4.493} & \textbf{5.776} & \textbf{5.134} & \underline{5.422} & \textbf{6.881} & \textbf{6.152} & \underline{5.360} & \textbf{6.079} & \underline{5.720} & \textbf{5.669} \\
 & SMOPD (token gate) & 4.471 & \underline{5.749} & 5.110 & 5.408 & \underline{6.876} & 6.142 & 5.355 & \textbf{6.079} & 5.717 & \underline{5.656} \\
\midrule
 \multirow{3}{*}{Qwen2.5-7B} & SMOPD (uniform) & 4.475 & 5.656 & 5.065 & \underline{5.448} & \underline{6.837} & \underline{6.143} & \textbf{5.421} & \underline{6.036} & \textbf{5.729} & \underline{5.646} \\
 & SMOPD (seq gate) & \textbf{4.497} & \underline{5.664} & \underline{5.080} & 5.447 & 6.826 & 6.136 & 5.383 & 5.991 & 5.687 & 5.635 \\
 & SMOPD (token gate) & \underline{4.495} & \textbf{5.669} & \textbf{5.082} & \textbf{5.452} & \textbf{6.847} & \textbf{6.149} & \underline{5.402} & \textbf{6.044} & \underline{5.723} & \textbf{5.651} \\
\midrule
 \multirow{3}{*}{Llama-3.2-3B} & SMOPD (uniform) & \textbf{4.437} & \textbf{5.633} & \textbf{5.035} & \textbf{5.349} & \textbf{6.756} & \textbf{6.052} & 5.406 & \underline{5.962} & \underline{5.684} & \textbf{5.590} \\
 & SMOPD (seq gate) & 4.413 & 5.588 & 5.001 & \underline{5.337} & 6.731 & \underline{6.034} & \textbf{5.427} & \textbf{5.974} & \textbf{5.701} & \underline{5.578} \\
 & SMOPD (token gate) & \underline{4.422} & \underline{5.611} & \underline{5.016} & 5.329 & \underline{6.738} & \underline{6.034} & \underline{5.416} & 5.919 & 5.667 & 5.572 \\
\bottomrule
\end{tabular}
\caption{Teacher-mixing comparison on safe alignment. Rows are SMOPD with uniform mixing (our main method) and the two confidence-failure gates. Per column and per backbone, \textbf{bold} marks the best and \underline{underline} the second best among the three mixing rules (ties share the rank).}
\label{tab:gate_safe}
\end{table*}

\section{API-Bank LLM-Judge Protocol and Original Exact-Match Analysis}
\label{app:apibank}

\begin{figure*}[!t]
\centering
\includegraphics[width=0.99\textwidth]{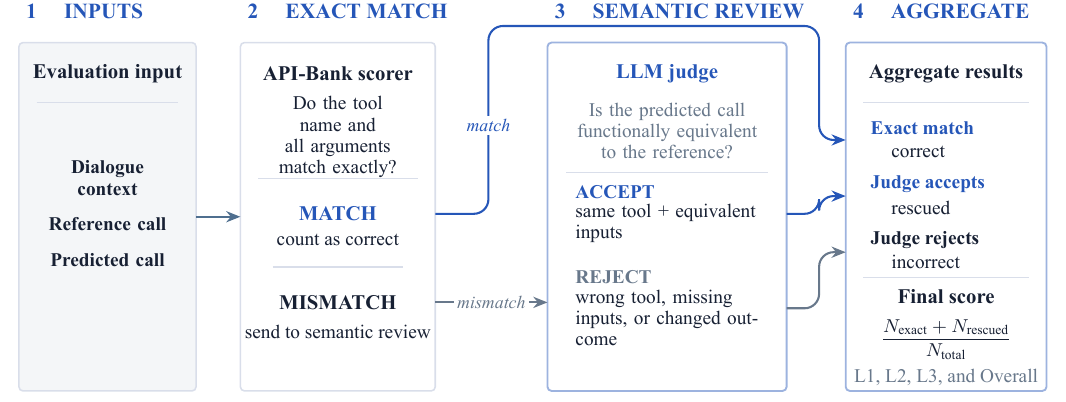}
\caption{API-Bank LLM-judge evaluation pipeline. The original exact matcher is applied first, and its successes remain correct. Only failed exact matches undergo semantic review using the saved dialogue, reference call, and prediction. A semantically equivalent call is rescued; a functionally different call remains incorrect. The final per-level and Overall scores combine original exact successes with judge-rescued failures.}
\label{fig:apibank_judge_pipeline}
\end{figure*}

The main text reports API-Bank under a semantic LLM-judge metric rather than the benchmark's original strict matcher. Figure~\ref{fig:apibank_judge_pipeline} summarizes the complete evaluation path, and Figure~\ref{fig:apibank_judge_prompt} shows the verbatim judge prompt. For every saved model response, we first apply the original exact matcher. Exact successes are retained as correct; only exact failures are sent to the judge (\texttt{qwen3.7-plus}, one query per failed item), together with the recent dialogue turns, the gold tool call, and the predicted call. The judge accepts a failure only when it uses the correct tool and its arguments are functionally equivalent to the requested call. Benign casing or formatting differences and harmless optional parameters may therefore be rescued, whereas wrong tools, missing required fields, unsupported values, or outcome-changing arguments remain incorrect. No model is re-run.

The final metric is $(N_{\mathrm{exact}}+N_{\mathrm{rescued}})/N_{\mathrm{total}}$, computed separately for L1/L2/L3 and as micro-accuracy over all 597 items. Under this metric (Table~\ref{tab:acc_format}), SMOPD obtains the highest Overall score at both scales: $87.10\%$ at 1.5B and $77.39\%$ at 3B, respectively $3.52$ and $1.01$ percentage points above the strongest scalarized baseline.

\begin{figure*}[!t]
\centering
\setlength{\fboxrule}{0.6pt}
\fbox{\begin{minipage}{\dimexpr\textwidth-2\fboxsep-2\fboxrule\relax}\small
You are a strict evaluator of tool-calling correctness for the API-Bank benchmark.

You are given a multi-turn dialogue, the GROUND-TRUTH tool call that correctly solves the user's final need, and a MODEL's predicted tool call. Decide whether the model's prediction is FUNCTIONALLY CORRECT: it must call the right tool AND supply parameters that faithfully fulfill the user's request as expressed in the dialogue, i.e.\ be semantically equivalent to the ground truth.

Judge CORRECT (Yes) when the prediction differs from the ground truth ONLY in ways that do not change the outcome, for example:
\begin{itemize}[nosep,leftmargin=1.4em]
\item Case, whitespace, or punctuation differences in parameter values.
\item Synonyms / equivalent phrasings that refer to the same entity or value.
\item Reasonable interpretations of time boundaries that still capture the user's stated intent (e.g.\ ``up to March 12th'' as 2023-03-12 23:59:59 vs 2023-03-12 00:00:00).
\item Equivalent formatting of the same value (e.g.\ 5 vs ``5'', equivalent date formats).
\item Extra optional parameters that are consistent with the dialogue and do not alter the core action.
\end{itemize}

Judge INCORRECT (No) when the prediction:
\begin{itemize}[nosep,leftmargin=1.4em]
\item Calls a different or wrong tool, or one that does not fulfill the request.
\item Omits a required parameter, or fills a required parameter with a wrong or contradictory value.
\item Uses a value that changes the meaning or outcome (different user, amount, target, time span, etc.).
\item Hallucinates values not supported by the dialogue.
\end{itemize}

Only the FINAL required tool call matters. Base your decision on the user's actual need in the dialogue, not on superficial string matching. Be strict: do not pass genuinely wrong calls, but do not fail calls that are merely phrased or formatted differently.

\#\ Dialogue (most recent turns)\\
\texttt{\{dialogue\}}

\#\ Ground-truth tool call\\
\texttt{\{gold\}}

\#\ Model predicted tool call(s)\\
\texttt{\{pred\}}

Respond in EXACTLY this format and nothing else:\\
Reasoning: \textless one concise sentence\textgreater\\
Judgment: \textless Yes or No\textgreater
\end{minipage}}
\caption{Verbatim prompt of the API-Bank LLM judge (\texttt{qwen3.7-plus}). The placeholders \texttt{\{dialogue\}}, \texttt{\{gold\}}, and \texttt{\{pred\}} are filled with the recent dialogue turns, the reference tool call, and the model's predicted call(s) of each exact-match failure; the judge returns a one-sentence rationale and a binary verdict.}
\label{fig:apibank_judge_prompt}
\end{figure*}

\begin{table*}[!t]
\centering
\small
\setlength{\tabcolsep}{2pt}
\begin{tabular}{@{}llll@{}}
\toprule
\textbf{Item} & \textbf{Reference param} & \textbf{3B output param} & \textbf{Difference} \\
\midrule
L1-8  & \texttt{symptom:"fatigue"}       & \texttt{symptom:"Fatigue"}                     & casing \\
L1-9  & \texttt{symptom:"fatigue"}       & \texttt{symptom:"Chronic fatigue syndrome"}    & more specific \\
L1-30 & \texttt{content:"Meeting"}       & \texttt{content:"Meeting with team"}           & fuller context \\
L1-32 & \texttt{appointment\_id:"78901234"} & \texttt{...,\ date:"2023-08-05"}            & extra field \\
\bottomrule
\end{tabular}
\caption{Representative API-Bank items where the 3B model calls the correct tool but is scored wrong by exact name-and-parameter matching. Each 3B output is semantically correct or more complete than the reference; the failure is purely a literal-string mismatch.}
\label{tab:apibank_cases}
\end{table*}

\paragraph{Original exact-match metric.}
The original API-Bank scorer requires exact equality of both the function name and the \emph{entire} parameter dictionary; any difference in casing, specificity, formatting, or additional fields counts as a failure. Table~\ref{tab:apibank_exact} preserves these strict scores for comparison with the original ToolRL/GD$^2$PO evaluation protocol. Unlike the LLM-judge results in the main text, they primarily measure literal agreement with the reference call.

\begin{table*}[!t]
\centering
\small
\setlength{\tabcolsep}{1.5pt}
\begin{minipage}[t]{0.49\textwidth}
\centering
\begin{tabular}{@{}l@{\hspace{5pt}}r@{\hspace{3pt}}r@{\hspace{3pt}}r@{\hspace{5pt}}r@{}}
\toprule
\textbf{Method} & \textbf{L1} & \textbf{L2} & \textbf{L3} & \textbf{Overall} \\
\midrule
\multicolumn{5}{l}{\textbf{Qwen2.5-1.5B}} \\
GRPO Baseline & 62.66 & 53.73 & \textbf{48.85} & 58.63 \\
GDPO Baseline & \underline{67.42} & \underline{56.72} & \underline{47.33} & \underline{61.81} \\
GD$^2$PO Baseline & 64.66 & 53.73 & 45.04 & 59.13 \\
\cmidrule(l){1-5}
Accuracy Teacher [0.9,0.1] & 64.91 & 58.21 & 49.62 & 60.80 \\
Format Teacher [0.1,0.9] & 72.18 & 62.69 & 38.17 & 63.65 \\
\cmidrule(l){1-5}
SMOPD & \textbf{74.44} & \textbf{64.18} & 39.69 & \textbf{65.66} \\
\bottomrule
\end{tabular}
\end{minipage}
\hfill
\begin{minipage}[t]{0.49\textwidth}
\centering
\begin{tabular}{@{}l@{\hspace{5pt}}r@{\hspace{3pt}}r@{\hspace{3pt}}r@{\hspace{5pt}}r@{}}
\toprule
\textbf{Method} & \textbf{L1} & \textbf{L2} & \textbf{L3} & \textbf{Overall} \\
\midrule
\multicolumn{5}{l}{\textbf{Qwen2.5-3B}} \\
GRPO Baseline & 66.67 & 47.76 & 27.48 & 55.95 \\
GDPO Baseline & 65.41 & \textbf{56.72} & \underline{38.17} & \underline{58.46} \\
GD$^2$PO Baseline & \underline{66.92} & \underline{50.75} & \textbf{41.98} & \textbf{59.63} \\
\cmidrule(l){1-5}
Accuracy Teacher [0.9,0.1] & 67.67 & 52.24 & 30.53 & 57.79 \\
Format Teacher [0.1,0.9] & 68.67 & 52.24 & 42.75 & 61.14 \\
\cmidrule(l){1-5}
SMOPD & \textbf{67.92} & 49.25 & 32.82 & 58.12 \\
\bottomrule
\end{tabular}
\end{minipage}
\caption{Original strict API-Bank exact-match accuracy (\%) on the same saved generations as Table~\ref{tab:acc_format}. \textbf{Overall} is micro-accuracy over all 597 items (399 L1, 67 L2, and 131 L3). These are the original name-and-full-parameter-dictionary matching scores, not the semantic LLM-judge metric used in the main text. Per column and backbone, \textbf{bold} marks the best and \underline{underline} the second best among the three scalarized baselines and SMOPD; single-reward teachers are shown for context and excluded from ranking.}
\label{tab:apibank_exact}
\end{table*}

\paragraph{Quantitative diagnosis of strict matching.}
On the 597 items shared by the 1.5B and 3B SMOPD runs, the exact matcher counts 392 correct for 1.5B and 347 for 3B. Of the 66 items that 1.5B passes but 3B fails, \textbf{62} are cases where 3B invokes the \emph{correct} tool but its argument string does not match the reference literally, and only 4 are genuine refusals or clarification requests. These disagreements concentrate on the simplest Level-1 items (38 of 66), precisely where the target is a short literal string and extra model capability cannot help.
The lower exact-match score of Qwen2.5-3B is therefore a literal-matching artifact rather than a regression in tool-calling ability: the more capable backbone tends to produce richer argument strings---more specific values, fuller context, or extra optional fields (Table~\ref{tab:apibank_cases})---which deviate from the short reference strings more often, so the strict matcher penalizes exactly the additional capability that the semantic judge rescues.

\paragraph{Representative cases.}
Table~\ref{tab:apibank_cases} lists 3B outputs that are semantically correct---often \emph{more} complete or faithful to the user's request---yet scored wrong by exact matching.

We use the semantic judge in the main table because it measures functional equivalence rather than literal reproduction, while retaining Table~\ref{tab:apibank_exact} for direct comparability with prior exact-match reporting. Together with BFCL, the judge results show that strict matching can understate functional tool-use ability without changing the central within-backbone comparison: at both scales, SMOPD attains the highest judged API-Bank accuracy of all compared methods.

\begin{table*}[t]
\centering
\small
\setlength{\tabcolsep}{1.2pt}
\begin{tabular}{llccccccccccccccc}
\toprule
& & \textbf{AST} & \multicolumn{5}{c}{\textbf{Non-Live AST}} & \multicolumn{5}{c}{\textbf{Live AST}} & \textbf{Multi} & \multicolumn{2}{c}{\textbf{Relevance}} & \textbf{Over-} \\
\cmidrule(lr){4-8}\cmidrule(lr){9-13}\cmidrule(lr){15-16}
\textbf{Size} & \textbf{Method} & \textbf{(ours)} & Acc & Simp & Mult & Par & P.M. & Acc & Simp & Mult & Par & P.M. & \textbf{Turn} & Rel & Irrel & \textbf{all} \\
\midrule
\multirow{8}{*}{1.5B}
& GRPO Base & 68.6 & 76.90 & 67.58 & 83.50 & 82.00 & 74.50 & 60.33 & 70.16 & 58.21 & 56.25 & 50.00 & 0.00 & 100.00 & 35.93 & 17.32 \\
& GDPO Base & 72.6 & 80.60 & 70.92 & 88.00 & 85.00 & 78.50 & 64.54 & 74.81 & 62.49 & 56.25 & 50.00 & 0.00 & 100.00 & 35.27 & 18.04 \\
& GD$^2$PO Base & 73.3 & 80.35 & 71.92 & 88.50 & 82.00 & 79.00 & 66.32 & 74.81 & 64.58 & 56.25 & 58.33 & 0.00 & 100.00 & 35.15 & 18.18 \\
& Accuracy Teacher & 70.5 & 78.83 & 70.83 & 87.50 & 81.00 & 76.00 & 62.10 & 72.09 & 59.73 & 56.25 & 62.50 & 0.00 & 100.00 & 37.44 & 17.84 \\
& Format Teacher & 70.5 & 76.92 & 69.17 & 85.50 & 79.50 & 73.50 & 64.03 & 73.26 & 62.01 & 56.25 & 58.33 & 0.00 & 100.00 & 15.72 & 15.67 \\
& SMOPD & 70.7 & 78.31 & 71.25 & 88.50 & 79.00 & 74.50 & 63.06 & 72.09 & 61.06 & 56.25 & 58.33 & 0.00 & 93.75 & 23.65 & 16.50 \\
& seq gate & 70.9 & 78.23 & 70.92 & 88.50 & 78.00 & 75.50 & 63.58 & 71.32 & 61.92 & 56.25 & 58.33 & 0.00 & 100.00 & 21.02 & 16.28 \\
& token gate & 70.9 & 78.08 & 71.33 & 88.50 & 78.00 & 74.50 & 63.66 & 74.03 & 61.54 & 50.00 & 54.17 & 0.00 & 100.00 & 20.02 & 16.18 \\
\midrule
\multirow{8}{*}{3B}
& GRPO Base & 75.8 & 83.00 & 73.50 & 91.50 & 85.00 & 82.00 & 68.54 & 63.18 & 70.28 & 68.75 & 50.00 & 0.00 & 81.25 & 79.50 & 23.10 \\
& GDPO Base & 75.3 & 80.56 & 71.25 & 91.00 & 82.50 & 77.50 & 70.02 & 65.89 & 71.04 & 75.00 & 66.67 & 0.00 & 87.50 & 76.80 & 22.74 \\
& GD$^2$PO Base & 76.5 & 82.69 & 73.75 & 92.50 & 84.00 & 80.50 & 70.32 & 67.83 & 71.04 & 75.00 & 62.50 & 0.00 & 81.25 & 75.57 & 22.86 \\
& Accuracy Teacher & 77.2 & 82.83 & 74.33 & 92.00 & 81.50 & 83.50 & 71.58 & 69.77 & 72.17 & 75.00 & 62.50 & 0.00 & 87.50 & 62.78 & 21.72 \\
& Format Teacher & 76.1 & 83.15 & 74.08 & 92.00 & 84.50 & 82.00 & 69.06 & 64.34 & 70.47 & 75.00 & 54.17 & 0.00 & 81.25 & 76.86 & 22.91 \\
& SMOPD & 77.2 & 83.15 & 73.58 & 92.50 & 84.50 & 82.00 & 71.21 & 68.60 & 72.08 & 75.00 & 58.33 & 0.00 & 87.50 & 73.42 & 22.78 \\
& seq gate & 77.3 & 83.71 & 73.83 & 93.00 & 84.50 & 83.50 & 70.84 & 66.67 & 72.17 & 68.75 & 58.33 & 0.00 & 93.75 & 73.23 & 22.78 \\
& token gate & 76.2 & 82.52 & 73.58 & 92.50 & 82.50 & 81.50 & 69.95 & 66.67 & 71.13 & 62.50 & 58.33 & 0.00 & 87.50 & 74.93 & 22.74 \\
\bottomrule
\end{tabular}
\caption{Full per-category BFCL-v4 accuracy (\%) for the aggressive-profile models of Table~\ref{tab:acc_format}. \textbf{AST (ours)} is the mean of the \emph{Non-Live} and \emph{Live AST} block accuracies (our main-text BFCL metric); P.M.\ abbreviates Parallel-Multiple. The full-suite \textbf{Overall} average is dominated by categories these single-turn tool callers never see (Multi-Turn; Web-Search and Memory return no score and are omitted), motivating the AST metric used in the main text.}
\label{tab:bfcl_full_ext}
\end{table*}

\section{Full BFCL-v4 Category Breakdown}
\label{app:bfcl}

Table~\ref{tab:bfcl_full_ext} reports every BFCL-v4 category for the aggressive-profile models of Table~\ref{tab:acc_format}. It shows why the full-suite \emph{Overall} score (last column, 15--23\%) is misleadingly low: it averages in categories our single-turn tool models were never trained for---\emph{Multi-Turn} is $0\%$ for every model, and the Web-Search/Memory categories return no score---while the models are in fact strong on the abstract-syntax-tree (AST) function-calling categories they target (76--84\% non-live, 60--72\% live). Our main-text \textbf{BFCL AST} metric is the mean of the Non-Live and Live AST columns. At 3B every SMOPD variant matches or exceeds the GDPO baseline's AST accuracy, whereas at 1.5B the scalarized baselines remain strongest (GD$^2$PO $73.3\%$ vs.\ SMOPD $70.7\%$).

\end{document}